\documentclass[letterpaper]{article} 

\usepackage{aaai24}  
\usepackage{times}  
\usepackage{helvet}  
\usepackage{courier}  
\usepackage[hyphens]{url}  
\usepackage{graphicx} 
\usepackage{natbib}  
\usepackage{caption} 
\usepackage{subcaption}

\usepackage{algorithm}
\usepackage{algorithmic}
\usepackage[usenames,dvipsnames]{xcolor}
\usepackage{amssymb}
\usepackage{amsmath}
\usepackage{mathtools}
\usepackage{cleveref}
\usepackage{tikz}
\usetikzlibrary{bayesnet}
\usetikzlibrary{arrows}
\usepackage{booktabs}

\usepackage{newfloat}
\usepackage{listings}
\usepackage{dsfont}
\usepackage{nicefrac}
\DeclareCaptionStyle{ruled}{labelfont=normalfont,labelsep=colon,strut=off} 
\floatstyle{ruled}
\newfloat{listing}{tb}{lst}{}
\floatname{listing}{Listing}
\def\Algname{\textcolor{black}{\textbf{I}mage \textbf{C}lassification \textbf{E}xplanations tailored to User \textbf{E}xpertise}}
\def\Algnameabbr{{\color{black} I-CEE}}
\def\Algtitle{\color{black} Image Classification Explanations tailored to User Expertise}
\def\query{{\color{black} Hypercorrection Effect}}

\DeclareMathOperator*{\argmax}{argmax} 

\newcommand{\summation}[2]{\sum\limits^{#1}_{#2}}

\title{I-CEE: Tailoring Explanations of Image Classification Models to User Expertise}

\author{
    Yao Rong\textsuperscript{\rm 1},
    Peizhu Qian\textsuperscript{\rm 2},
    Vaibhav Unhelkar\textsuperscript{\rm 2},
    Enkelejda Kasneci\textsuperscript{\rm 1}
}
\affiliations{
    \textsuperscript{\rm 1}Technical University of Munich \\ \textsuperscript{\rm 2}Rice University\\
    \{yao.rong, enkelejda.kasneci\}@tum.de, \{pqian, vaibhav.unhelkar\}@rice.edu

}

\begin{document}
\maketitle

\begin{abstract}

Effectively explaining decisions of black-box machine learning models is critical to responsible deployment of AI systems that rely on them.
Recognizing their importance, the field of explainable AI (XAI) provides several techniques to generate these explanations.
Yet, there is relatively little emphasis on the user (the explainee) in this growing body of work and most XAI techniques generate ``one-size-fits-all'' explanations.
To bridge this gap and achieve a step closer towards human-centered XAI, we present \Algnameabbr{}, a framework that provides \Algname{}.
Informed by existing work, \Algnameabbr{} explains the decisions of image classification models by providing the user with an informative subset of training data (i.e., example images), corresponding local explanations, and model decisions.
However, unlike prior work, \Algnameabbr{} models the \textit{informativeness} of the example images to depend on user expertise, resulting in different examples for different users.
We posit that by tailoring the example set to user expertise, \Algnameabbr{} can better facilitate users' understanding and simulatability of the model.
To evaluate our approach, we conduct detailed experiments in both simulation and with human participants ($N = 100$) on multiple datasets.
Experiments with simulated users show that \Algnameabbr{} improves users' ability to accurately predict the model's decisions (simulatability) compared to baselines, providing promising preliminary results.
Experiments with human participants demonstrate that our method significantly improves user simulatability accuracy, highlighting the importance of human-centered XAI.
\end{abstract}

\section{Introduction}
As AI systems receive increasingly important roles in our life, human users are challenged to comprehend the decisions made by these systems.
To ensure user safety and proper use of AI systems, experts across disciplines have recognized the need for AI transparency~\cite{yang2017evaluating, ehsan2021expanding, russell2021human}.
Solutions for AI transparency -- e.g., techniques for explainable AI (XAI) -- are essential as most AI models can be viewed as a ``black box,'' whose decision-making process cannot be easily interpreted or understood by human users. 
Among the different settings of XAI, our work focuses on explaining image classification tasks~\cite{arrieta2020xai}.
Existing XAI techniques for image classification widely use attribution explanations, such as GradCAM~\cite{selvaraju2017grad}, LIME~\cite{ribeiro2016should}, SHAP~\cite{lundberg2017unified} and its extended methods~\cite{wang2022accelerating,chuang2023cortx}. 
While these techniques inform our work, they all miss one key element: human factors, potentially due to the complexity of modeling human users. 

\begin{figure}[t]
    \centering
    \includegraphics[width=.95\linewidth]{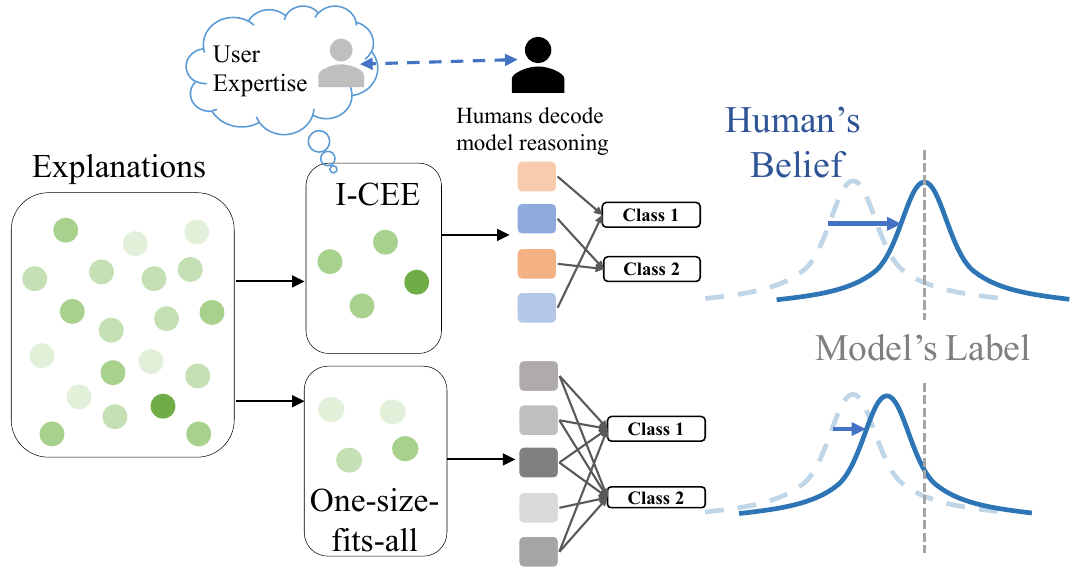}
    \caption{\Algnameabbr{} tailors the explanation process to each user by considering their expertise. By selecting the most informative explanations based on user expertise, \Algnameabbr{} can better enhance user simulatability of ML model's decisions.}
    \label{fig:teaser figure}
\end{figure}

We advocate that human modeling is critical to XAI research because explainability is inherently centered around humans~\cite{liao2021human}.
A few works focusing on explaining reinforcement learning policies use cognitive science theories to model the human user and generate explanations based on the human model \cite{baker2011bayesian, huang2019enabling, lage2019exploring, qian2022evaluating}.
Closer to our focus, the works of \citet{yang2022psychological} and \citet{yang2021mitigating} utilize a Bayesian Teaching framework to model human perception and then generate human-centered explanations.
One limitation of these works is that all human users are treated the same by the modeling method, presuming that an identical set of explanations will work for \textit{all} users.
In contrast, we attempt to generate tailored explanations for each user by modeling their \textit{task-specific expertise}. 
Our approach to modeling user expertise is informed by human annotator models used in active and imitation learning~\cite{welinder2010multidimensional,beliaev2022imitation}.
Similar to these works, our user model aims to capture both the decisions and reasoning process (expertise in concepts used for image classification) of the human user in the context of a given classification task.

To bridge the research gap that personalization is missing in the explanation process, we propose the framework \Algname{} (\Algnameabbr{}).
Informed by existing XAI methods for image classification, our framework utilizes the \textit{explanation-by-examples} paradigm and provides attribution explanations (local explanations) for a subset of training data.
However, in \Algnameabbr{}, the approach of selecting the example explanations differs and is user-specific. 
For a given image classification task, \Algnameabbr{} first discovers a set of $m$ task-relevant concepts. It then models the user's task-specific expertise as a $m$-dimensional vector, where each entry lies between $[0,1]$ and represents their expertise in the corresponding concept. Based on this user model, \Algnameabbr{} finally selects the set of local explanations that can best fill user's knowledge gaps.

As depicted in \Cref{fig:teaser figure}, by selecting the set of local explanations that can best increase the user's task-specific expertise, \Algnameabbr{} aims to accelerate user's understanding of the decision-making process of the machine learning model.
In contrast, most existing work in XAI either selects random or one-size-fits-all local explanations, thereby foregoing the opportunity to accelerate model understanding by providing tailored explanations.
The contributions of this work can be summarized as follows:
\begin{itemize}
    \item We identify the opportunity for tailored explanations for explaining decisions made by image classification models and develop a novel framework named \Algnameabbr{} that realize this opportunity. This work represents an advancement towards human-centered explanations.
    \item To evaluate \Algnameabbr{}, we test the simulatability of explanations generated by our framework on four datasets. Results demonstrate that our framework achieves better simulatability (i.e., users' ability to predict the model's decisions) relative to state-of-the-art XAI baselines\footnote{Code is available at \url{https://github.com/yaorong0921/I-CEE}.}.
    \item We evaluate our framework through detailed human-subject studies ($N = 100$). Experimental results indicate that our framework can more effectively help users understand the ML model's decision-making than the state-of-the-art technique Bayesian Teaching~\cite{yang2021mitigating}, and is subjectively more preferred by the participants, highlighting the advantages of our framework.
\end{itemize}

\begin{figure*}[t]
    \centering
    \includegraphics[width=.95\linewidth]{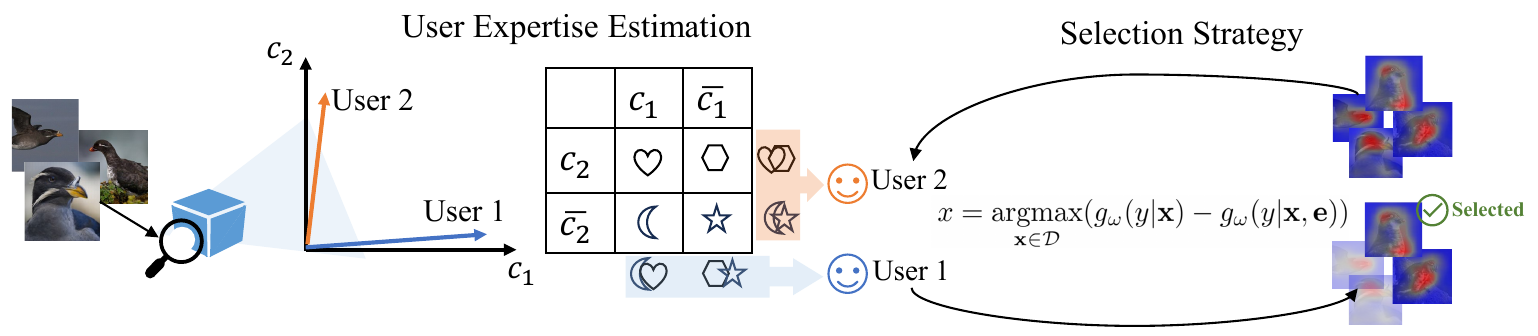}
    \caption{Overview of \Algnameabbr{}. \textbf{Left:} The target model is first projected into a concept space, which is then used to estimate user expertise. Two users are illustrated. User 1 uses the concept $c_1$ in the reasoning process and can differentiate only two classes (highlighted in blue). Likewise, User 2 is able to distinguish two classes based on $c_2$ (in orange). \textbf{Right:} Based on user models, explanations with images $(\mathbf{x}, \mathbf{e})$ in the training set that maximize \query{} are selected and delivered to the users.}
    \label{fig:method}
\end{figure*}

\section{Related Work}
\paragraph{Human-centered Explainable AI.} 
Recent surveys indicate a growing activity in XAI research~\cite{doshi2017towards,liao2021human,rong2023towards, chuang2023efficient}.
The field recognizes the central role of humans in their explanations, leading to increasing adoption of human-centered evaluations of explanation techniques~\cite{lage2019human}.
Besides evaluations, a few techniques have also considered human factors in generating explanations~\cite{lage2020learning, lage2019exploring, huang2019enabling, qian2022evaluating, yang2022psychological}.
Among these, the most related framework is that of Bayesian Teaching, which focuses on image classification and selects explanations by modeling the users as a Bayesian agent~\cite{yang2021mitigating}.
However, this work does not model differences between users' reasoning or prior expertise.
In contrast, we consider personalized user models to better fit the specific explanation needs of different users.
Our design is informed by research in pedagogy and active machine learning.

\paragraph{Pedagogical Theories on Learning from Errors.} 
XAI has been viewed as a teaching process, where the XAI technique serves the role of the teacher and the user that of the student~\cite{qian2022evaluating}.
To teach learners effectively, pedagogical research confirms that a teacher needs to assess a learner's prior knowledge and design instructions accordingly \cite{owens2017teaching,ambrose2018learning}. 
A common indicator of incorrect knowledge is errors, caused by an incorrect association or understanding.
To correct the errors, feedback on the correct answers along with explanations have been found to be crucial and most helpful \cite{metcalfe2017learning}. 
These findings in learning sciences have laid the groundwork for our XAI framework, motivating our example selection approach; in particular, \Algnameabbr{} emphasizes explaining the images on which it estimates the user will make errors.
Additionally, as the confidence in an error increases, learning from the error also increases \cite{butterfield2001errors, metcalfe2011hypercorrection}.
This is an effect known as the hypercorrection effect. 
To reflect the hypercorrection effect in our framework, we choose images where the user has low confidence in the correct label (i.e., high confidence in the incorrect label), and argue that using these examples will result in better learning outcomes. 

\paragraph{Active Learning.}
In the context of machine learning (ML), techniques for active learning aim to achieve high model accuracy while minimizing the required labeling effort~\cite{settles2009active, ren2021survey}.
Active learning is valuable in domains where a limited amount of training data is labeled, and it has been used beyond classification tasks such as in sequence labeling~\cite{settles2008analysis} or image semantic segmentation~\cite{sinha2019variational}.
While active learning pertains to training machines, we observe that insights from the field are highly relevant for XAI (which seeks to train humans about an AI model).
By making this novel connection, we leverage a central component of active learning techniques -- \textit{query strategies} -- to inform the development and evaluation of \Algnameabbr{}.

\section{Problem Statement}
Consider an ML classifier, denoted as $f$ or the \textit{target model}, trained on dataset $\mathcal{D}$ of image-label pairs $(\mathbf{x},y)$.
The classifier $f : \mathbb{R}^d \rightarrow \{1:K\}$ maps an input image $\mathbf{x} \in \mathbb{R}^d$ to a label $y \in \{1:K\}$, i.e., $f(\mathbf{x}) = y$, where $K$ is the number of classes.
For a subset of images, the predicted label $y$ may not match the true label $y^*$.
To explain such target models, different feature attribution methods have been proposed that generate local explanations~\cite{ribeiro2016should, lundberg2017unified}.
These local explanation assigns each input pixel an importance value, denoted as $\mathbf{e} \in \mathbb{R}^d$, which is usually visualized as a saliency map.
In the \textit{explanation-by-example} paradigm, the user is shown a set of images sampled from the training data, its local explanation, and its prediction, i.e., $(\mathbf{x}, \mathbf{e}, y)$. 
As the user has limited time to understand the model, it is important to select the set of most informative example images.

Within the explanation-by-example paradigm, we consider the problem of selecting the set of most informative example images (and corresponding explanations).
Formally, our problem assumes three inputs: the target model $f$, a data set $\mathcal{D}$ ($|\mathcal{D}|=N$), and a feature attribution method to generate local explanations.
Given these inputs, we seek to generate a subset $S \subset \mathcal{D}$ of training data composed of $M\ll N$ images that best facilitate \textit{simulatability}, i.e., help users predict the decisions of the ML model.
As the problem objective hinges on a human-centered metric, its successful resolution warrants a human-centered approach.

\section{\Algnameabbr{}: \Algtitle}
We now present our approach to solve this problem: \Algnameabbr{}, which is composed of two phases (\Cref{fig:method}).
First, our framework models the user by estimating their task-specific expertise (lines 3-4, \Cref{alg:icee}).
Second, by simulating the user using this model and a query strategy, \Algnameabbr{} selects informative example images and explanations (lines 5-8).

\subsection{User Expertise Estimation}
The process of a user predicting an ML model's labeling decisions can be viewed as one of image annotation, where the annotators might possess distinct areas of strengths or \textit{expertise} affecting their giving labels~\cite{welinder2010multidimensional}. 
For instance, some users find textual patterns to be more recognizable than shapes while others find shapes to be more intuitive.
During the annotation process, humans frequently use ``concept-based thinking'' in reasoning and decision making: identifying similarities among various examples and organizing them systematically based on their resemblances~\cite{yeh2020completeness,armstrong1983some,tenenbaum1999bayesian}.
Recognizing these aspects of human reasoning and informed by annotator models proposed in active learning, we model a user by estimating their expertise in applying different task-relevant concepts. 
We first discover the underlying concepts in the feature space of the target model.
Using the discovered concepts, we model a user with a vector representing their ability to utilize each concept when annotating images.

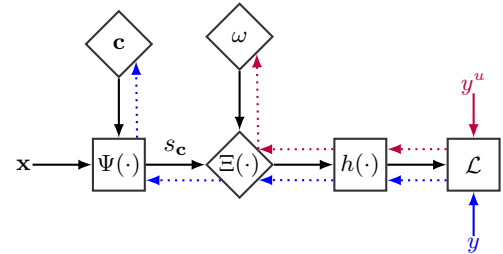
\begin{figure}[t]
\resizebox{.8\linewidth}{!}{
    \begin{tikzpicture}
    \tikzstyle{main}=[rectangle, minimum size = 7mm, thick, draw =black!80, node distance = 8mm]
    \tikzstyle{connect}=[-latex, thick]
    \tikzstyle{connectr}=[-latex, thick, draw=purple]
    \tikzstyle{connectb}=[-latex, thick, draw=blue]
    \tikzstyle{connectrdot}=[-latex, thick, dotted, draw=purple, auto shift]
    \tikzstyle{connectbdot}=[-latex, thick, dotted, draw=blue, auto shift,]
    \tikzstyle{box}=[diamond, minimum size = 9mm, thick, draw =black!80, node distance = 8mm]
     \node[] (x) [label=center:$\mathbf{x}$] { };
      \node[main, fill = white!100] (sc) [right=of x, label=center:$\Psi(\cdot)$] { };
      \node[box] (psi) [right=of sc,label=center:$\Xi(\cdot)$] { };
      \node[box] (w) [above=of psi,label=center:$\omega$] { };
      \node[main] (h) [right=of psi,label=center:$h(\cdot)$] {};
      \node[box] (c) [above=of sc,label=center:$\mathbf{c}$] { };
      \node[main] (l) [right=of h,label=center:$\mathcal{L}$] { };
      \node[] (y) [below=6mm of l, label=center:\textcolor{blue}{$y$}] { };
      \node[] (y2) [above=6mm of l, label=center:\textcolor{purple}{$y^u$}] { };
      \path (x) edge [connect] (sc)
            (sc) edge [connect] node[above] {$s_{\mathbf{c}}$} (psi)
            (psi) edge [connect] (h)
    		(c) edge [connect] (sc)
    		(w) edge [connect] (psi)
    		(h) edge [connect] (l)
              (y) edge [connectb] (l)
              (y2) edge [connectr] (l)
              (l.150) edge [-latex, thick, dotted, draw=purple, auto=right]  (h.30)
              (h.150) edge [-latex, thick, dotted, draw=purple, auto=right]  (psi.39)
              (psi.40) edge [-latex, thick, dotted, draw=purple, auto=right]  (w.-45)
              (l.-150) edge [-latex, thick, dotted, draw=blue, auto=right]  (h.-30)
              (h.-150) edge [-latex, thick, dotted, draw=blue, auto=right]  (psi.-39)
              (psi.-140) edge [-latex, thick, dotted, draw=blue, auto=right]  (sc.-30)
              (sc.56) edge [-latex, thick, dotted, draw=blue, auto=right]  (c.-45);
    \end{tikzpicture}
    }
    \caption{User Modeling: Square nodes are deterministic, while diamond nodes are trainable. Loss back-propagated for concept discovery (Eq.~\ref{eq:loss}) is marked in blue, while that for expertise estimation (Eq.~\ref{eq:omega}) is in red.}
    \label{fig:user model}
\end{figure}
\Cref{fig:user model} provides an overview of the user model.
To arrive at the model, \Algnameabbr{} begins with applying the concept discovery algorithm on the target model~\cite{yeh2020completeness} that aims to recover $m$ concept $[\mathbf{c}_1, \cdots \mathbf{c}_m]$ , such that 
\begin{equation}
    f(\mathbf{x}) = h(\Psi(\mathbf{x})) = h(\Xi_\theta(s_\mathbf{c}(\mathbf{x})))
\end{equation}
where $\Psi(\mathbf{x}) \equiv [\psi(\mathbf{x}^1), \dots, \psi(\mathbf{x}^T)]$ are $T$ activation vectors,
$h(\cdot)$ represents the mapping from the intermediate output of activation vectors to image labels,\footnote{$\Psi$ and $h$ can also be viewed as the intermediate and final layers of the image classification neural network, respectively. As $h$ and $\Psi$ are not trained as part of the user model, we do not explicitly denote their parameters (such as weights and biases) in our notation.}
$s_\mathbf{c}(\cdot)$ is the concept score
\begin{equation}
    s_\mathbf{c}(\mathbf{x})= \langle \psi(\mathbf{x}^i), \mathbf{c}_j\rangle|_{j=1}^m |_{i=1}^T\in \mathbb{R}^{m \cdot T}
\end{equation}
that estimates the alignment between each concept and activation vector pair, and
$\Xi_\theta: \mathbb{R}^{T\cdot m} \rightarrow \mathbb{R}^{T\cdot n}$ is a trainable mapping that converts concept scores back into the activation space.
Both the concept vectors and concept scores are unit normalized.
For concept discovery (i.e., computing $\mathbf{c}, \theta$), the following cross-entropy loss is minimized:
\begin{equation}
    \mathcal{L}_{(\mathbf{c}, \theta)} = - \summation{N}{i=1} y_i\log(h(\Xi_\theta(s_\mathbf{c}(\mathbf{x}_i)))),
\label{eq:loss}
\end{equation}
where $y$ is the prediction from the target model $f(\cdot)$.

After completing concept discovery (which is a one-time process), the expertise estimation for each user takes place within the concept space.
We freeze all model parameters ($\Psi(\cdot)$, $s_{\mathbf{c}}(\cdot)$, $\Xi_\theta(\cdot)$ and $h(\cdot)$) trained using Eq.~\ref{eq:loss} to learn an expertise vector $\omega \in \mathbb{R}^m$ for each user.
The variations among users are manifested through different values of $\omega$, as their diverse domain knowledge influences the way they utilize concepts to arrive at predictions.
Concretely, we ask users to annotate images and use $\omega$ to simulate their predictions.
The expertise vector ${\omega}$ for a user is learned by minimizing the following cross-entropy loss:
\begin{equation}
    \mathcal{L}_\omega = - \summation{N}{i=1} y^u_i\log(h(\Xi_\theta(\omega \cdot s_{\mathbf{c}}(\mathbf{x}_i))), \label{eq:omega}
\end{equation}%
where $y^u$ denotes annotated labels collected from the user.
Once $\omega$ is learned, we obtain a user model denoted as $g_{\omega}(\cdot) = h(\Xi_\theta(\omega \cdot s_{\mathbf{c}}(\cdot))$.
If $\omega_1 \approx \omega_2$, it implies that these two users (Users 1 and 2) have very similar ``reasoning process" as the utilization of concepts is very similar.
Likewise, if $\omega \approx \mathbf{1}_{m}$, this user employs a very similar reasoning mechanism as the target model $f$.

\begin{algorithm}[t]
   \caption{\Algnameabbr{}}
   \label{alg:icee}
  \begin{algorithmic}[1]
   \STATE {\bfseries Input:} Target model $f(\cdot)$, data $\mathcal{D}$, user annotation $y^u$. 
   \STATE {\bfseries Output:} A set of example images and explanations $\mathcal{S}$.
   \STATE Discover concepts by solving Eq.~\ref{eq:loss}.
   \STATE Estimate user expertise by solving Eq.~\ref{eq:omega}.
   \FOR{$\mathbf{x} \in \mathcal{D}$}{
    \STATE {Calculate \query{} for $\mathbf{x}$ using Eq.~\ref{eq:query}.}
   }
   \ENDFOR
   \STATE Return top-$K$ image samples.
\end{algorithmic}
\end{algorithm}

\subsection{Selection Strategy}
Our goal is to select a set of informative examples that can most improve the user's simulatability.
To estimate the informativeness of the examples, we employ the concept of the hypercorrection effect in educational psychology. As the human needs to learn how the model makes the decision, the model's prediction is viewed as the ``correct" answer whereas the human's disagreed initial belief is the ``error". Feedback on the correct answer along with explanations has been found to be crucial and most helpful in learning new knowledge~\cite{metcalfe2017learning}. As the confidence in an error increases, i.e., the confidence in the correct answer decreases, learning from this error example is more effective~\cite{butterfield2001errors, metcalfe2011hypercorrection}. To reflect the hypercorrection effect in \Algnameabbr{}, we choose images where the user has lower confidence in the model's predicted label after knowing the model's reasoning and argue that using these examples will lead to higher learning outcomes. 
Concretely, \Algnameabbr{} aims to identify a set of examples $\mathcal{S} \subseteq \mathcal{D}$ which consists of samples with the top maximal \query:
\begin{equation}
        x = \argmax_{\mathbf{x} \in \mathcal{D}} \underbrace{(g_{\omega}(y|\mathbf{x})-g_{\omega}(y|\mathbf{x}, \mathbf{e}))}_{\text{\query{} of $\mathbf{e}$}},
\label{eq:query}
\end{equation}
where $g_{{\omega}}(\cdot)$ represents the user model, $\mathcal{D}$ denotes the training dataset, and $\mathbf{e}$ and $y$ are the local explanation and machine prediction corresponding to the image $\mathbf{x}$.

\section{Experiments with Simulated Users}
Before conducting a user study, we first evaluate our approach through extensive experiments with simulated users on one synthetic and three realistic image classification tasks.
To facilitate reproducibility, Appendix includes more details about the experimental setup.

\begin{figure}[t]
     \begin{subfigure}[b]{0.49\linewidth}
         \centering
         \includegraphics[width=\textwidth]{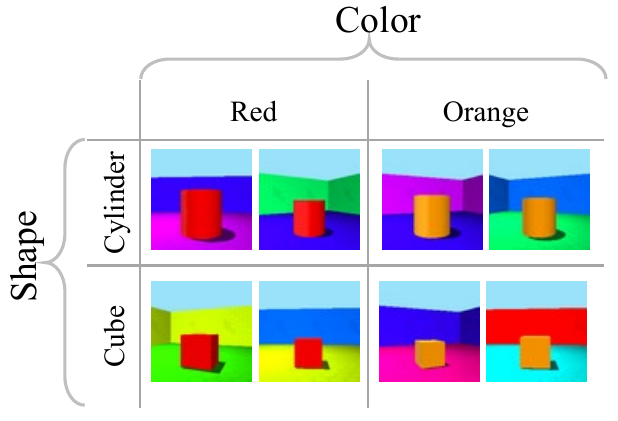}
         \caption{Synthetic dataset}
         \label{fig:synthetic dataset}
     \end{subfigure}
     \hfill   
     \begin{subfigure}[b]{0.45\linewidth}
         \centering
         \includegraphics[width=\textwidth]{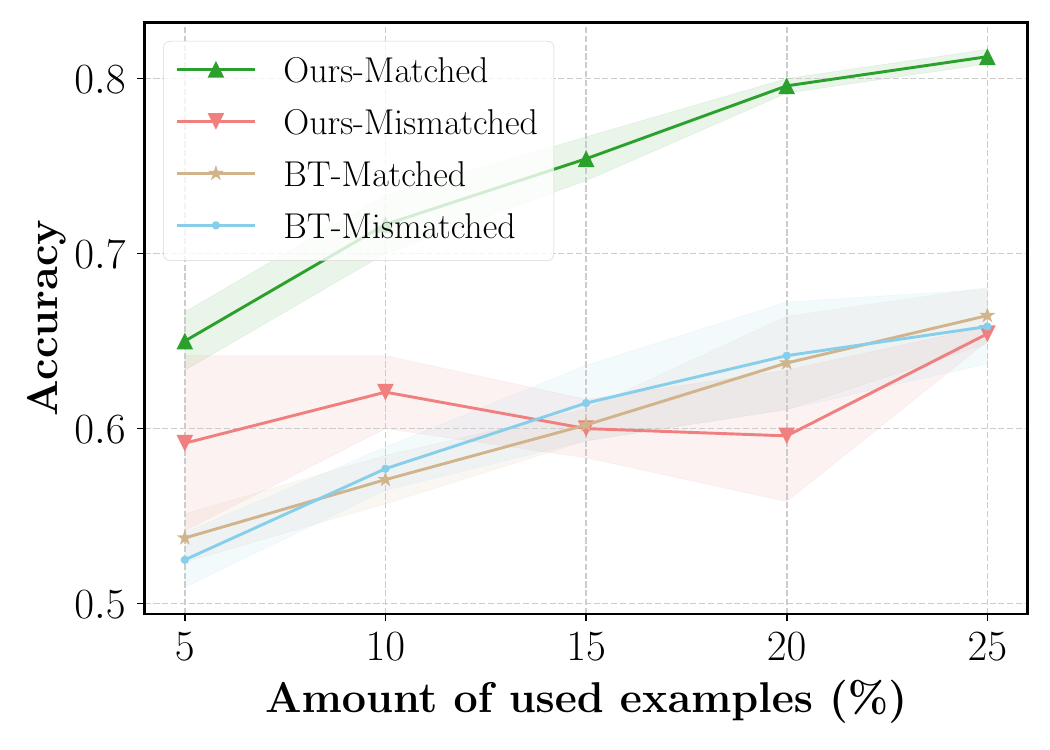}
         \caption{(Mis-)Matched expertise}
         \label{fig:match expertise}
     \end{subfigure}
     \caption{\textbf{(a)}: Overview of four classes in the synthetic dataset. \textbf{(b)}: User simulatability accuracy when trained with examples that match/mismatch with the user expertise. }
\end{figure}

\begin{figure*}[t]
     \centering
     \begin{subfigure}[b]{0.26\textwidth}
         \centering
         \includegraphics[width=\textwidth]{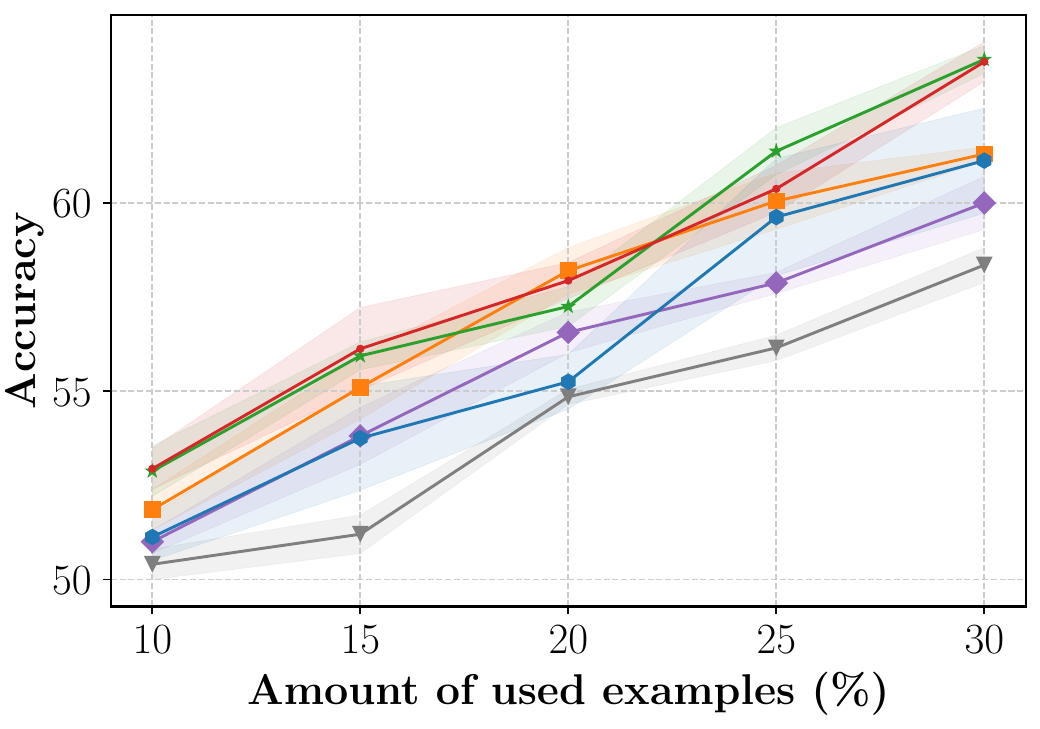}
         \caption{CIFAR-100}
         \label{fig:cifar}
     \end{subfigure}
     \hfill
     \begin{subfigure}[b]{0.26\textwidth}
         \centering
         \includegraphics[width=\textwidth]{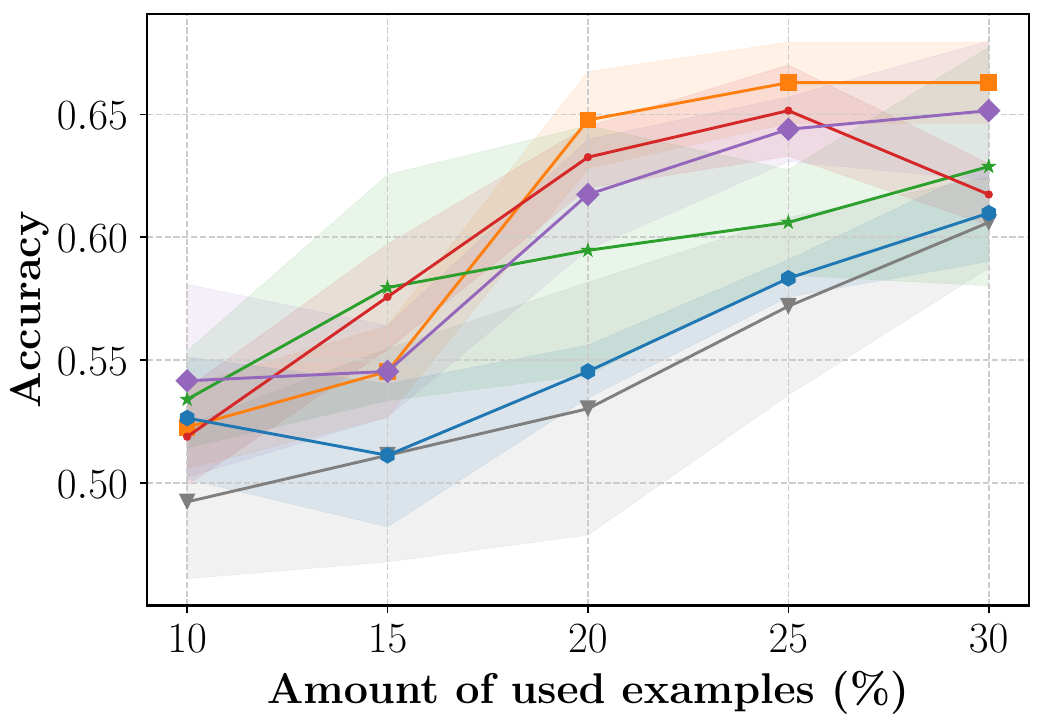}
         \caption{CUB-200-2011}
         \label{fig:cub}
     \end{subfigure}
     \hfill
     \begin{subfigure}[b]{0.36\textwidth}
         \centering
         \includegraphics[width=\textwidth]{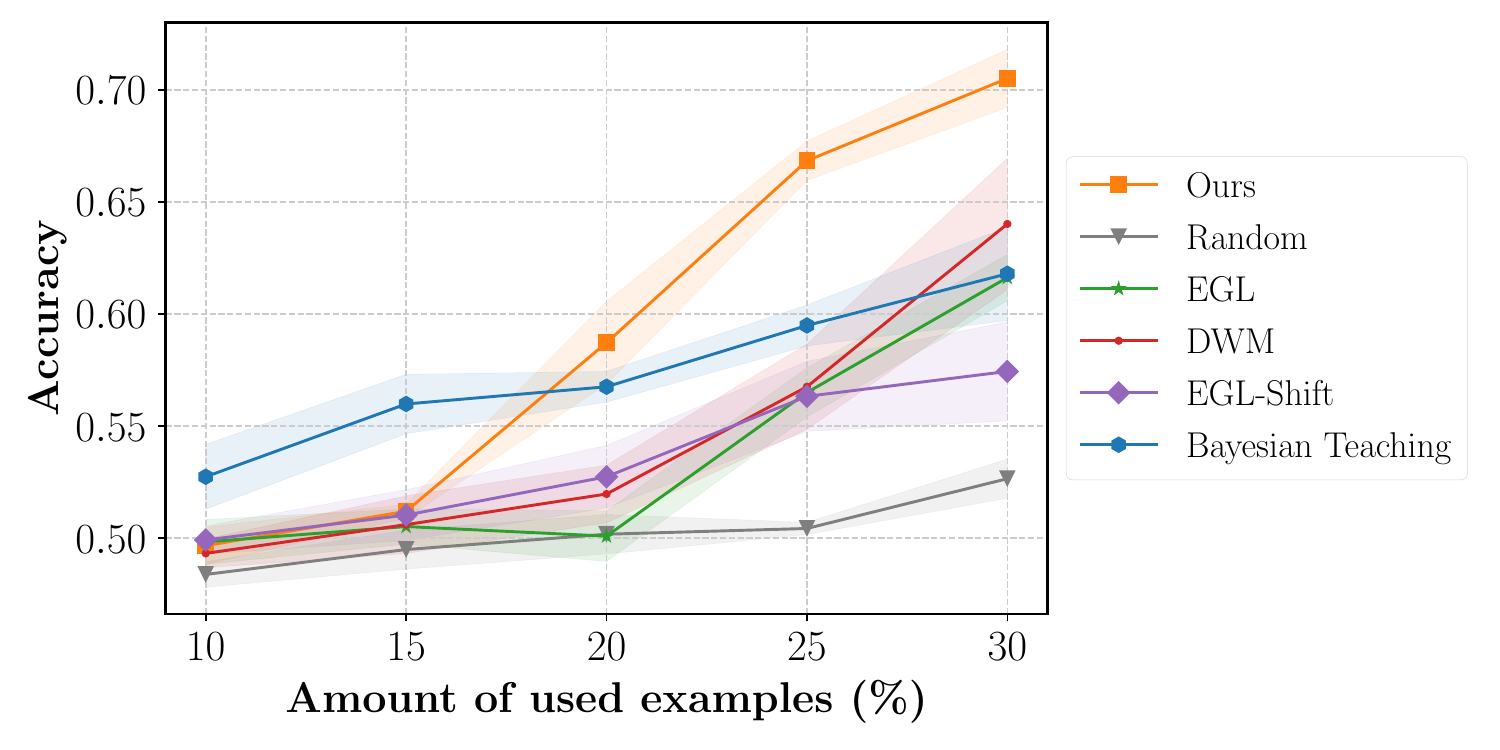}
         \caption{GTSRB}
         \label{fig:gtsrb}
     \end{subfigure}
        \caption{Comparison with baseline algorithms with simulated users on three datasets. The ratio of used examples $p$ (in percentage) is plotted on the x-axis and simulatability accuracy is on the y-axis. (Results averaged over $5$ runs.)}
        \label{fig:comparison}
\end{figure*}
\paragraph{Synthetic Dataset.}
We construct a synthetic dataset\footnote{This dataset is based on 3d-shapes~\cite{kim2018disentangling}.} to validate the design of our proposed method in simulation. 
This dataset contains four classes and each class is described with two concepts, color and shape, illustrated in \Cref{fig:synthetic dataset}. 
For instance, if a user uses colors to distinguish between different classes (i.e., they have more expertise in using ``colors" then ``shapes"), then to this user, the red cylinders and red cubes belong to the same class, which differs from the orange ones. 
Likewise, for a user who has high expertise in using shapes, the cylinders and the cubes are distinguishable for this user regardless of their colors. 
The other visual features such as angles or background colors are randomly sampled as they are not essential in this decision-making process. 
For each class, we generate $300$ images ($80\%$ for training and $20\%$ for testing). 
We use a ResNet-18~\cite{he2016deep} as our classification model and use GradCAM~\cite{selvaraju2017grad} for generating explanations. 
Given their annotation behavior, a simulated behavior is modeled using Eqs.~\ref{eq:loss}-\ref{eq:omega}, i.e., identical to the modeling approach of \Algnameabbr{.}

\paragraph{Realistic Datasets.}
We also benchmark \Algnameabbr{} on three real-world datasets: CIFAR-100~\cite{krizhevsky2009learning}, CUB-200-2011~\cite{wah2011caltech} and German Traffic Sign Recognition Benchmark (GTSRB)~\cite{stallkamp2012man}. 
We construct a simulated user from pre-defined annotations on each dataset who behaves differently from the target model. 
In particular, for each dataset, our simulated user can distinguish only two classes out of four similar classes.
All methods are evaluated based on this user. 
For instance, on CUB-200-2011, the simulated user labels both Crested and Least Auklet as the same class (Crested Auklet), and Parakeet and Rhinoceros Auklet as the same class (Parakeet Auklet). 
We use the original training-test splits on these datasets and, similar to the procedure in the synthetic dataset, we use ResNet-50~\cite{he2016deep} for classification training and GradCAM for computing explanations.

\subsection{Baseline Methods}
We evaluate \Algnameabbr{} against a recent human-centered XAI approach: Bayesian Teaching (BT) ~\cite{yang2021mitigating}.
BT simulates a user's behavior (i.e., their prediction of an image class) by deploying a ResNet-50-PLDA (probabilistic linear discriminate analysis ~\cite{ioffe2006probabilistic}) model.
By assuming users perform Bayesian reasoning, it selects example images and explanations to better align user's beliefs to the target model.
\Algnameabbr{} and BT differ in their approaches to both user modeling and example selection.

To evaluate the example selection alone, we also benchmark against query strategies derived from active learning (AL).
Unlike traditional AL, in our application of AL query strategies to XAI, the simulated user is the learner and the target model is the annotator.
We use Expected Gradient Length (EGL)~\cite{settles2007multiple}, Density-Weighted Method (DWM)~\cite{settles2008active} as well as a random sampling strategy as baselines. 
EGL, in the context of this paper, selects samples $(x,e)$ that result in the greatest change to the current model if the annotated label is known. The ``change" imparted to the model from the queried samples is measured by the gradient of the objective function with respect to the model parameters. 
However, the instances chosen by EGL might be outliers that cause significant gradient changes. To alleviate this issue, \citet{settles2008active} proposes to integrate a density-weighting technique with the query strategy such as EGL. 
Specifically, each sample is weighted with its average similarity to all other instances in the input dataset. 
In this work, we extend EGL with the belief shift in the calculated EGL when considering $e$ in the input (denoted as EGL-Shift). Specifically, we compute the difference between EGL of $(x,e)$ and $x$. With EGL-Shift, we aim to alleviate the influence of an image itself on the training gradient but emphasize the impact of explanations.

\subsection{Evaluation Metric}
To evaluate our method, we use simulatability, which is commonly used as a proxy for testing a user's understanding of the model's decision-making process~\cite{hase2020evaluating, arora2022explain, hase2020leakage}. 
Simulatability is measured as ``to what extent can a user successfully predict a model's prediction.'' 
This metric can be used in both simulation experiments and human user studies.

We follow the experimental settings proposed in ~\cite{yeh2018representer,koh2017understanding} to study the influence of selected examples. 
Specifically, each method provides an ordered set of example images $\mathcal{S}$, where the ranking is decided by the \textit{informativeness} defined in the respective method.
We denote the ratio between number of example images $|\mathcal{S}|$ and the size of training data $\mathcal{D}$ as $p = \nicefrac{|\mathcal{S}|}{|\mathcal{D}|}$.
The simulated user is retrained using these example images $\mathbf{x}$ and their corresponding labels $y = f(\mathbf{x})$, where recall that $f$ is the target model. 
Given the retrained user model $g_{\omega}'$, we compute the user's accuracy of predicting the model's predictions on the test set, i.e., the simulatability of the user:
\begin{equation}
    \text{Acc} = \frac{1}{N_t}\summation{N_t}{i=1} \mathds{1}(y_i = g_{\omega}'(\mathbf{x_i})), 
\end{equation}
where $N_t$ is the number of samples in the test set.

\begin{figure*}[t]
     \centering
     \begin{subfigure}[b]{0.32\linewidth}
         \centering
         \includegraphics[height=4.2cm]{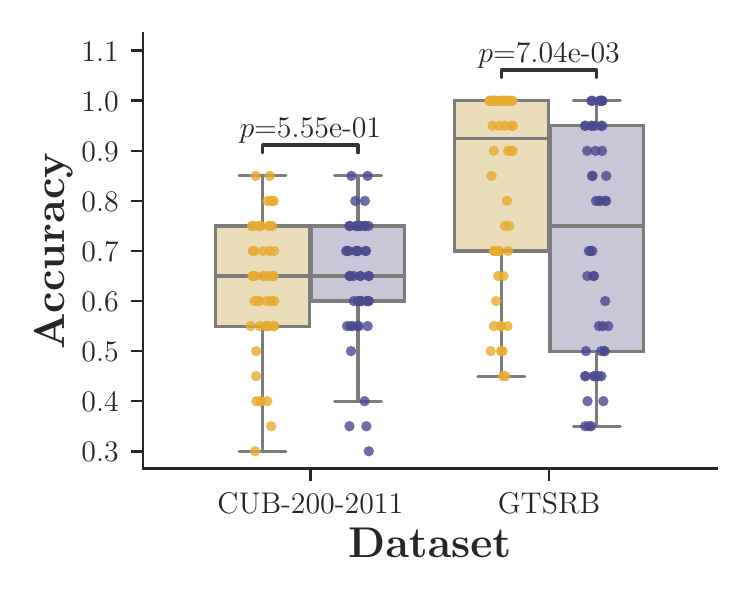}
         \caption{User Accuracy (all predictions)}
         \label{fig:user acc}
     \end{subfigure}
     \begin{subfigure}[b]{0.32\linewidth}
         \centering
         \includegraphics[height=4.2cm]{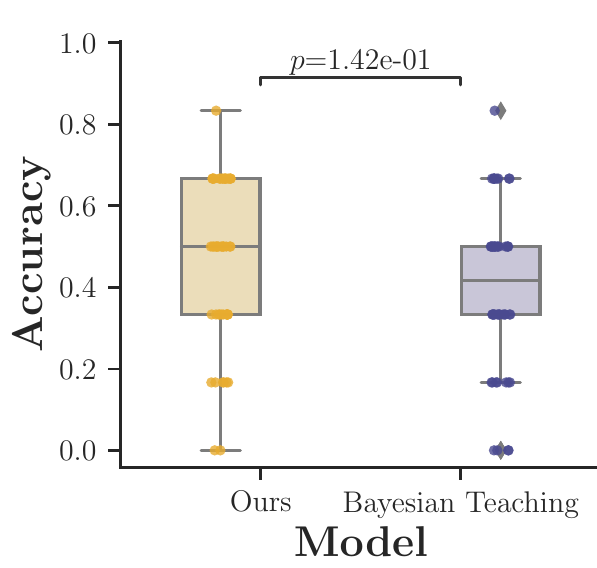}
         \caption{User Accuracy (inaccurate pred., CUB)}
         \label{fig:user acc cub}
     \end{subfigure}
     \begin{subfigure}[b]{0.35\linewidth}
         \includegraphics[height=4.2cm]{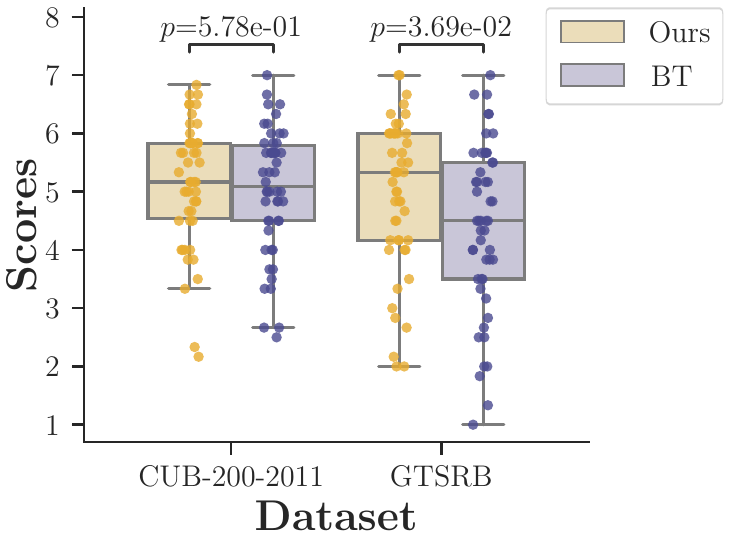}
         \caption{User evaluation score.}
         \label{fig: user score}
     \end{subfigure}
        \caption{Results of experiments with human users $(N=100)$ comparing \Algnameabbr{} with the baseline Bayesian Teaching (BT). (a) Simulatability accuracy on all predictions, (b) Simulatability accuracy on images where the target model made inaccurate predictions in the CUB-200-2011 dataset, (c) User's subjective perception of model explanations.}
        \label{fig:three graphs}
\end{figure*}

\subsection{Experimental Results}

\subsubsection{Ablation Study.}
To validate our model design of $g(\cdot)$, we study (1) whether $\omega$ can faithfully reflect the user expertise and 
(2) the advantages of tailored explanations according to the user expertise. 
We simulate two users on the synthetic dataset: User 1 only uses color in classification while User 2 only uses shape. 
We deduce annotations for each user based on attributes for each class (\Cref{fig:synthetic dataset}). 

After estimating each user, we investigate their expertise vector: $\omega_1$ and $\omega_2$ ($\omega_i \in \mathbb{R}^8$).
Each entry in $\omega_i$ represents the expertise of the user in one specific concept. 
The top four largest entries in $\omega_1$ and $\omega_2$ are complementary, corresponding to the fact that each user has the opposite expertise (i.e., each user uses different concepts in the decision-making).
To validate the efficacy of the user model via expertise, we run an experiment where we train User 1 using a set of examples specifically chosen based on the User 1 model (``Matched"), against a set of examples chosen for User 2 (``Mismatched").
As demonstrated in \Cref{fig:match expertise}, we observe that the simulated user achieves high simulatability accuracy when they receive examples selected according to their expertise (``Ours Matched"). 
However, if selecting examples that do not maximize the \query{} tailored to the particular user (``Ours Mismatched"), the simulatability accuracy is low, indicating that such examples fail to provide substantial insights into the target model.
Additionally, we compare our user simulation model to that of Bayesian Teaching.
We observe little differences between the matched and mismatched settings using the BT framework, suggesting that BT might not be able to accurately simulate the different behaviors of various users. Consequently, it cannot provide examples that effectively improve user simulatability (less performance improvement compared to ours).
\subsubsection{Comparison.}

We compare \Algnameabbr{} with baselines on three real-world datasets in \Cref{fig:comparison}.
Evaluation in user prediction accuracy is conducted at $p=[10, 15, 20, 25, 30]\%$. On CIFAR-100, our method always outperforms BT and EGL-Shift but is inferior to EGL and DWM. 
A potential reason for this result is that the explanation of CIFAR-100 is vague due to the low resolution of images. 
In this case, \query{} cannot be well captured since explanations are noisy. On CUB-200-2011 and GTSRB, our method outperforms other baselines at most of the percentages. 
For instance, on CUB our method achieves the best performance after $20\%$. Note that $20\%$ of the train data consists of $24$ images. 
This is a reasonable number of samples that can be efficiently studied by human users, which we will show in the next section. 
On GTSRB, we observe an evident performance gap between our method and the competitive baseline BT. 
A possible explanation for this can be attributed to the architecture of the user model: our model simulates the user via learning $\omega$ in the concept space without weakening the capability of the final classifier. 
On the contrary, BT relies on a PLDA layer to classify images, which can result in suboptimal performance when the latent features of images are highly similar, such as in traffic signs. 
This is not desirable because humans are good at distilling critical concepts and filtering out similar but irrelevant visual features. 
With more precise user modeling, our method demonstrates the capability of offering informative learning samples in most of the cases within the simulation experiments.

\section{Experiments with Human Users}
We conduct a human user study using the CUB-200-2011 and GTSRB datasets following the same settings as in the simulation experiments. We choose these two datasets as they are more challenging and the images are in higher resolution. 
We use Bayesian Teaching~\cite{yang2021mitigating} as a baseline since it is the most state-of-art and closest to our focus. 
Users are first asked to study two classes (among which there are actually four classes) and write down the features used to distinguish between these classes. This step is to let the user think as the pre-defined simulated user, to whom we have tailored model explanations. Then, $20$ model explanations selected by our method (experimental group) or Bayesian Teaching (control group) for users are shown, and we ask them to write down the features they use to determine the model prediction. 
During the evaluation section, participants first receive a test with $15$ questions to predict the model's label (images used here are sampled from the test set and include all four classes evenly). We refer to this section as ``objective understanding". Then, participants rate their perceived understanding on seven questions on the 7-Likert scale, which we refer to as ``subjective understanding". 
In the user study, we aim to study the following research questions:
\begin{itemize}
    \item \textbf{R1}: Our framework selects informative samples that can increase human understanding of the model.
    \item \textbf{R2}: Human understanding of the model is affected by task domains. 
\end{itemize}

\paragraph{Participants.}
We recruited $100$ participants (average age is $28.8 \pm 8.6$, $49$ females, $50$ males, and $1$ undefined) using a research platform Prolific\footnote{https://www.prolific.co/}, and randomly assigned them to one of the two conditions ($50$ participants/condition). $51$ participants have prior experience with AI from using Alexa, Siri, ChatGPT, or from ML-related courses. 
All participants passed the attention check during the user study. 
The study protocol has been approved by the Technical University of Munich IRB. 
At the beginning of the experiment session, we collected informed consent through Prolific. Each participant was compensated with a payment of \pounds$4.50$ for participation in the user study (within 30 minutes).

 \begin{figure}[t]
     \centering
     \includegraphics[width=.97\linewidth]{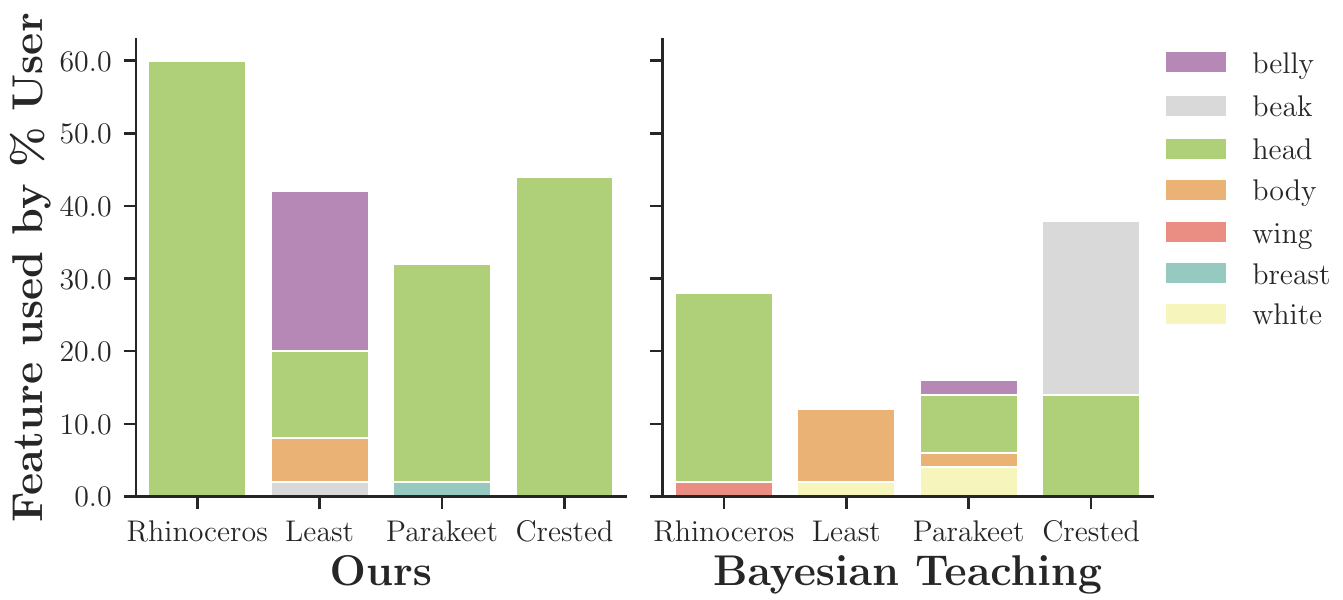}
     \caption{Illustration of features used by human users for distinguishing each class on CUB-200-2011.}
     \label{fig:user description}
 \end{figure}

\subsection{Results} 
\subsubsection{Analysis on \textbf{R1}.}
The results of the simulatability accuracy in each condition on each dataset are shown in \Cref{fig:user acc}. On GTSRB, we observe a statistically significant improvement in using our framework on user simulatability accuracy by $11.5\%$ ($p=0.007$). On the CUB dataset, we see that users from two conditions achieve similar user prediction accuracy and no significant effect is observed.
However, if we inspect the test samples where the target model makes inaccurate predictions (wrong classification) ($6$ out of $15$ images in the test are wrongly predicted), our method demonstrates superior performance compared to BT. 
Users from the experimental condition achieve an accuracy of $46.3\%$, whereas users from the control condition achieve $40.3\%$, as plotted in \Cref{fig:user acc cub}. 
These results indicate that users exhibit improved capability in simulating inaccurate predictions from the target model using our method, which is a more challenging task.
Additional evidence of the enhancement achieved through our model can be found in~\Cref{fig:user description}. We count the words of the features that users think the model uses to distinguish four different classes. When using our framework, the users tend to agree on the same feature (body part of the bird) for each class. For instance, about $68\%$ of the users use ``Head" to distinguish Rhinoceros, and about $20\%$ of the users think highly of ``Belly" for Least Auklet. Nevertheless, it is more difficult for users in Bayesian Teaching to come to an agreement, for example, for Least Auklet, only around $10\%$ of the participants use ``Body" as a feature while other users give diverse descriptions. 
These results highlight the advantage of the method in improving user understanding of the given target model.

As shown in \Cref{fig: user score}, the improvement in subjective understanding (rating scores) is not significant on CUB (average rating score is $5.14$ in our method and $5.02$ in BT). However, we observe that on GTSRB our method surpasses BT significantly with $p=0.037$.
The reason for significant improvement in GTSRB is that our method selects explanations bringing knowledge for distinguishing four classes. But BT chooses examples that reflect important features only for two classes, which hinders users from understanding how the model makes predictions for the other classes.

\subsubsection{Analysis on \textbf{R2}.}
The quantitative result shows that the task domain (dataset) affects the user's objective understanding. However, different tasks influence less subjective understanding, e.g., no significant difference between two datasets when using our method as illustrated in \Cref{fig: user score}. At the end of the user study, we asked participants for feedback on comparing the perceived helpfulness of model explanations in two datasets. While most of the users in both conditions find the explanations useful, seven users in the experimental condition and fourteen users in the control condition find the explanations on bird species are more helpful than the explanations on road signs. One reason causing this uncertainty in the road sign images is that the salient area is always a circle that covers the road sign, which seems to ``be the only one characteristic" for different classes.

\section{Conclusion}
We present a human-centered XAI framework, \Algnameabbr{}, that provides explanations of image classification ML models that are tailored to user expertise. 
Our framework first discovers task-relevant concepts, uses these concepts to arrive at expertise-based user models, and then selects examples and explanations that help the users to learn the missing concepts so they can accurately predict the machine's image classification decisions. 
We evaluate our approach through simulation experiments on four datasets, and report on a detailed human-subject study ($N = 100$). 
In these experiments, we observe that \Algnameabbr{} outperforms prior art, shows the promise of human-centered XAI, and motivates future research direction for the design of XAI systems.

\paragraph{Limitations and Future Work.}
Future investigation of our framework can consider the following avenues. 
First, more complex models of expertise estimation should be studied. 
In this work, we simulate user expertise by employing the concept-based reasoning approach for image classification proposed in~\cite{yeh2020completeness}. 
An alternative approach involves utilizing Large Language Models to simulate multiple humans in textual format~\cite{argyle2023out, aher2023using}.
Second, the current framework does not consider the sample complexity associated with user expertise estimation.
Future work should investigate methods that estimate user expertise with a small number of real-user annotations.
Third, we encourage replication of our work to be tested with different datasets, as the power of explanations is dependent on the task domain.
Future work should evaluate on datasets that include a more diverse pool of examples, as suggested by some of the participants.

\paragraph{Implications for XAI Systems.}
This study highlights the importance of personalized XAI, within the explanation-by-example paradigm for image classification. Future work should investigate the potential of personalized XAI in other contexts.
We argue that user modeling is essential to provide explanations that target user-specific misunderstanding or confusion.
Future XAI systems should leverage and address individual users' preferences and confusion. This involves the development of human-in-the-loop systems, allowing users to actively participate in the process of generating explanations.

\section{Ethical Statement}
In this work, we attempt to put human users at the center of XAI design, with the aim of creating AI systems that can be interpreted by non-expert end users.
To safeguard user privacy and user rights, we have received approval from University IRB.  
We believe that only when AI becomes more accessible, acceptable, and usable, can we realize its full potential to empower the world around us. 

\bibliography{aaai24}

\newpage

\part*{Appendix}
\begin{figure*}[t]
     \centering
     \begin{subfigure}[b]{0.24\linewidth}
         \centering
         \includegraphics[height=.8\textwidth]{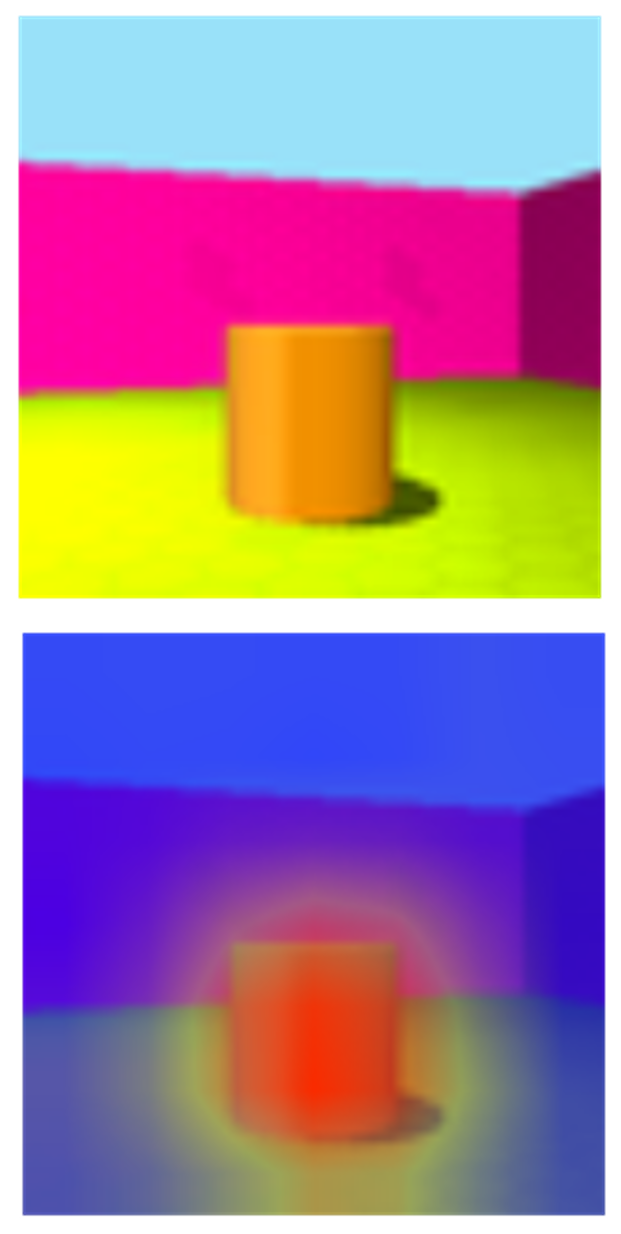}
         \caption{Synthetic dataset}
     \end{subfigure}
     \begin{subfigure}[b]{0.24\linewidth}
         \centering
         \includegraphics[height=.8\textwidth]{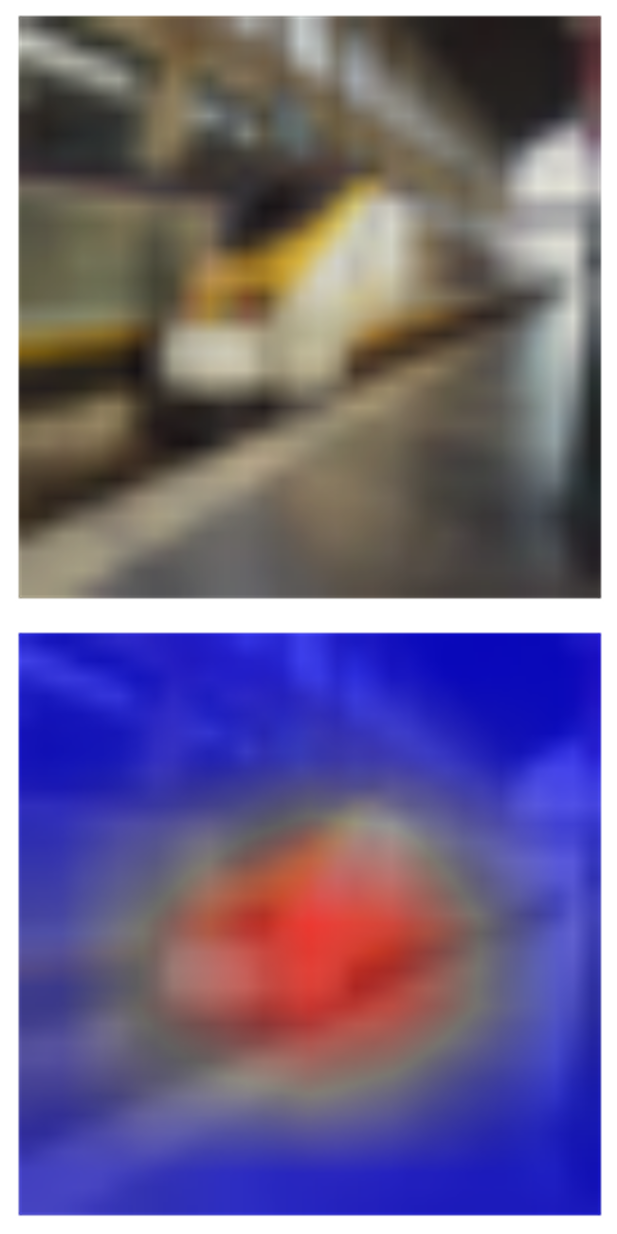}
         \caption{CIFAR-100}
     \end{subfigure}
     \begin{subfigure}[b]{0.24\linewidth}
         \centering
         \includegraphics[height=.8\textwidth]{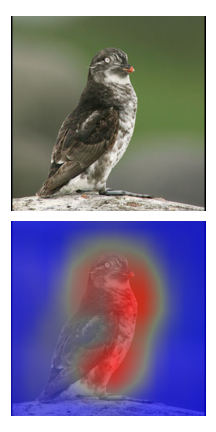}
         \caption{CUB-200-2011}
     \end{subfigure}
     \begin{subfigure}[b]{0.24\linewidth}
         \centering
         \includegraphics[height=.8\textwidth]{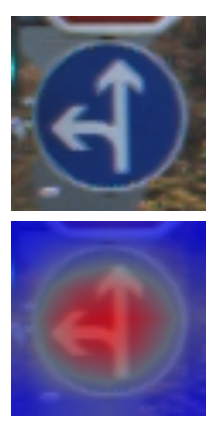}
         \caption{GTSRB}
     \end{subfigure}
        \caption{Illustration of model explanations on each dataset. The saliency map highlights the important area (feature) that is important for the model decision.}
        \label{fig:expl}
\end{figure*}

\begin{figure*}[t]
     \centering
     \begin{subfigure}[b]{0.24\linewidth}
         \centering
         \includegraphics[height=.6\textwidth]{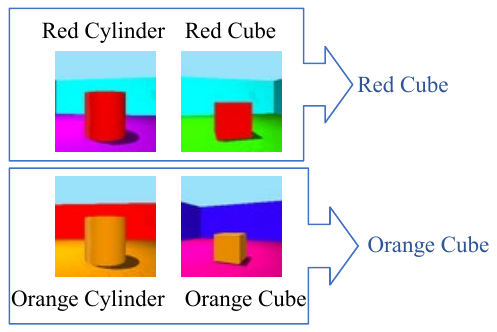}
         \caption{}
     \end{subfigure}
     \begin{subfigure}[b]{0.24\linewidth}
         \centering
         \includegraphics[height=.6\textwidth]{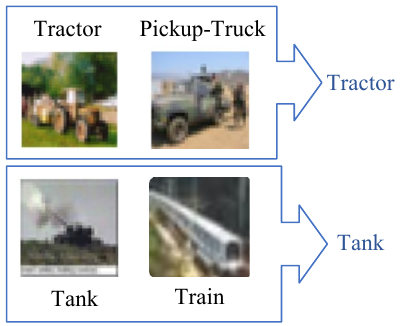}
         \caption{}
     \end{subfigure}
     \begin{subfigure}[b]{0.24\linewidth}
         \centering
         \includegraphics[height=.6\textwidth]{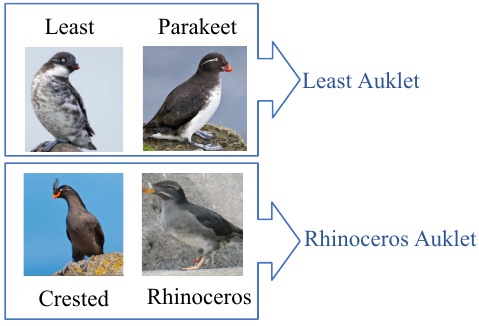}
         \caption{}
     \end{subfigure}
     \begin{subfigure}[b]{0.24\linewidth}
         \centering
         \includegraphics[height=.6\textwidth]{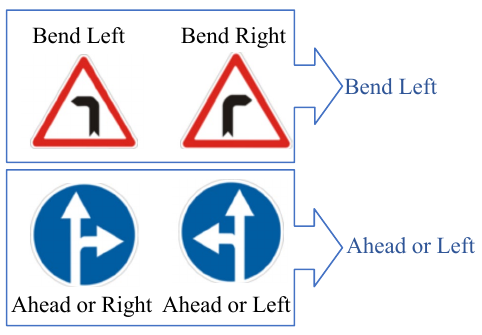}
         \caption{}
     \end{subfigure}
        \caption{Illustration of annotation given by the simulated user on the (a) synthetic, (b) CIFAR-100, (c) CUB-200-2011 and (d) GTSRB dataset. Original label is in black, and the label given by the simulated user is in blue.}
        \label{fig:user sim}
\end{figure*}
\section{Target Models and Explanations}
\subsection{Datasets}
We assess our method on four datasets. In the synthetic dataset, there are 960 images for training and 240 images for testing. These images are distributed across four classes (Red-Cylinder, Orange-Cylinder, Red-Cube, Orange-Cube). CIFAR-100~\cite{krizhevsky2009learning} comprises a total of 60,000 images, with 50,000 images designated for training and 10,000 images for testing. The dataset encompasses 100 diverse classes, each containing 600 images. CUB-200-2011 is a fine-grained dataset focusing on various bird species. The dataset encompasses a total of 11,788 images, distributed with 5,994 images for training and 5,794 images for testing. It comprises 200 different bird species, with an average of 30 images per species in training and 30 in testing.
CUB-200-2011~\cite{wah2011caltech} is a fine-grained dataset of bird species. The dataset consists of a total of 11,788 images, where 5,994 are for training and 5,794 for testing. It contains 200 different bird species, with each species having on average 30 images for training and 30 for testing. 
GTSRB (German Traffic Sign Recognition Benchmark)~\cite{stallkamp2012man} contains 51,840 images of German road signs in 43 classes. We use 80\% and 20\% of the whole dataset as training and testing sets, respectively.

\subsection{Target Model Details}
We finetune the ResNet-18~\cite{he2016deep} pre-trained on ImageNet as our target models on each dataset. On the synthetic dataset, the target models are trained using the Stochastic Gradient Descent (SGD) Optimizer with the learning rate of $1e^{-4}$ for 10 epochs. On realistic datasets, the learning rate is set to $1e^{-3}$ and the target model is trained for 50 epochs. Input images are resized to $224 \times 224$ on all datasets except CUB-200-2011, where images are resized to $448 \times 448$.
When training, random horizontal flipping is deployed as data augmentation. The first row in \Cref{tab:target model} lists the test accuracy of the target model on each dataset with all test classes. 

\begin{table}[h]
\centering
\resizebox{\linewidth}{!}{
\begin{tabular}{ccccccc}
\toprule[1pt]
           &Synthetic & CIFAR-100 & CUB-200-2011 & GTSRB \\\hline
Test  (all)     &    1.00       &    0.73          &     0.78        &        0.99  \\
Test  (subset)     &    1.00       &    0.82          &     0.81        &        0.99  \\
\bottomrule[1pt]
\end{tabular}
}
\caption{Accuracy of target models. The first row indicates the accuracy of all test classes. The second row contains the accuracy for classes selected for training simulated user models.}
\label{tab:target model}
\end{table}

\subsection{Target Model Explanations}
We employ GradCAM~\cite{selvaraju2017grad}, compute after the final convolutional block in the target model, as our chosen method for generating explanations. We choose this explanation is that GradCAM is very closed to human gaze-based attention in discovering distinguishable visual features~\cite{rong2021human}, which benefits human understanding compared to other explanation methods. 
The saliency map is resized to the original input, i.e.,  $\mathbf{x} \in \mathbb{R}^d$ and $\mathbf{e} \in \mathbb{R}^d$. \Cref{fig:expl} shows explanation examples on each dataset. We see for instance that the target model highlights the locomotive of the train on CIFAR-100, which represents the important feature of a train. Furthermore, on GTSRB, the ``left turn" on the sign is also highlighted by the saliency map. This study uses local explanations for their effectiveness in enhancing user understanding~\cite{rong2023towards}. Future work should explore more advanced model explanations, as our framework accommodates various explanations.
\section{Implementation Details of \Algnameabbr}
In this section, we provide an overview of the implementation details of our proposed method. This encompasses the training procedure for our simulated user model, as well as an explanation of the selection strategy we employ.
\subsection{User Model Training}
We select four very similar classes on each dataset to train simulated user models. We limit the number of studied classes because we aim to use them in studying the understanding of real human users based on the selected examples. The test performance of the target model on four selected classes is shown in the second row in \Cref{tab:target model}.
To obtain the simulated user, we first learn a concept in the latent space by using Eq.~(2) and Eq.~(3). On the synthetic dataset, we use $m=8$ using this dimension, the test accuracy reaches almost $100\%$, and thus no need to use a higher dimension number. 
On realistic datasets, the number of concepts $m$ can be set by end users. This is related to the dimension of the expertise vector $\omega \in \mathbb{R}^m$ (Eq.(4)). To choose a proper dimension of $\omega$, we train different concept spaces using different $m$ on each realistic dataset. More details can be found in the following section. The final setting is $m=64$ on each realistic dataset. 

After obtaining concepts, the user model $g_\omega(\cdot)$ is trained with the user annotations using Eq.(4). All other parameters in the network are frozen except $\omega$. To establish simulated users with specific expertise, we simulate user-annotated labels by mixing two classes into one class. This implies that the user must have different expertise from the target model because it cannot distinguish all four classes. The selected four classes and the user annotation on each realistic dataset are illustrated in \Cref{fig:user sim}. We train $g_\omega(\cdot)$ using Adam Optimizer with a learning rate of $1e^{-2}$ and for 40 epochs. Note that in experiments in Figure 5 in the main paper, the same training setting is used. 

\subsubsection{Hyper-parameter Settings.}
In this section, we show the choice of $m$, the number of concepts, on each realistic dataset. \Cref{tab:m} lists the accuracy of the trained simulated user (test with user annotations) and the number of trainable parameters in the concepts $\mathbf{c}$ as well as mapping function $\Xi(\cdot)$. The chosen $m$ is marked in bold according to the trade-off between the accuracy and the number of parameters. Lower number of trained parameters is preferred, if the user model needs to be deployed on resource-limited scenarios in real-world applications. From the results, we see that using $m=64$ on three datasets achieves already very high test accuracy. Incorporating additional concepts does not yield a substantial improvement in performance; instead, it introduces more computational costs.

\begin{table}[t]
\begin{subtable}[c]{0.5\textwidth}
\centering
\resizebox{\linewidth}{!}{
\begin{tabular}{c|c|c|c|c}
\toprule[1pt]
$m$       & 16 & 32 & \textbf{64} & 128  \\ \hline
Acc       &   85.00 $\pm$ 0.50 & 89.25 $\pm$ 0.34 &  93.50 $\pm$ 0.70  &  96.5$\pm$ 0.23 \\ \hline
\# Param. (M)&  0.94   & 0.97 &   1.05 &  1.19 \\ \bottomrule[1pt]
\end{tabular}
}
\subcaption{CIFAR-100}
\vspace{1em}
\end{subtable}
\begin{subtable}[c]{0.5\textwidth}
\centering
\resizebox{\linewidth}{!}{
\begin{tabular}{c|c|c|c|c}
\toprule[1pt]
$m$       &  16 & 32 & \textbf{64}  & 128\\ \hline
Acc       &  25.75 $\pm$ 0.78 & 27.27 $\pm$ 0.89 &  63.35 $\pm$ 0.45 &  65.15  $\pm$ 0.60    \\ \hline
\# Param. (M) & 3.67  &  3.73   & 3.85    &   4.08     \\ \bottomrule[1pt]
\end{tabular}
}
\subcaption{CUB-200-2011}
\vspace{1em}
\end{subtable}
\begin{subtable}[c]{0.5\textwidth}
\centering
\resizebox{\linewidth}{!}{
\begin{tabular}{c|c|c|c|c}
\toprule[1pt]
$m$       & 8 & 16 & 32 & \textbf{64}  \\ \hline
Acc       &  89.17$\pm$ 0.34 &   85.83  $\pm$ 0.35 &  98.33 $\pm$ 0.23  &    100.0  $\pm$ 0.10   \\ \hline
\# Param. (M) & 0.92  & 0.94   &  0.97   &   1.05      \\ \bottomrule[1pt]
\end{tabular}
}
\subcaption{GTSRB}
\vspace{1em}
\end{subtable}
\caption{Effect of $m$ on the user model performance.}
\label{tab:m}
\end{table}

\subsection{Selection Strategy}
We use Eq.(5) for selecting images that can better make users think more and learn more about the model reasoning mechanism from given examples. Based on the trained user model, we calculate the probability of the input image belonging to the class $y$ given by the target model, i.e., $g_\omega(y|\mathbf{x})$. When giving ($\mathbf{x},\mathbf{e}$) as the input, we deploy the explanation $\mathbf{e}$ as the weighted mask (the saliency map, achieved by normalizing the saliency map, onto the input. This approach is commonly used for evaluating the effectiveness of explanations~\cite{petsiuk2018rise,tomsett2020sanity}.


\section{Details of Baselines}
\subsection{Bayesian Teaching}
We implement Bayesian Teaching according to~\cite{yang2021mitigating}. We adapt their image selection strategy as we do not have two particular classes for our questions. In particular, the question used in \cite{yang2021mitigating} is a two-alternative forced choice task, where the authors use the Bayesian Teaching probability to choose two examples from the target model-predicted class and two examples from the alternative class (pre-defined by the authors). The selected images aim to lead the user model (explainee model) $f_L(\cdot)$ to classify a target image with the same label given by the target model. Therefore, we adapt their method by not choosing examples from the alternative class. 
Concretely, the probability we aim at is that $\mathbf{x}$ belongs to the class $y$ from which another image $\tau^y$ is sampled, which is denoted as $f(\mathbf{x}~|~\tau^{y})$ (borrowed from~\cite{yang2021mitigating}).
Under the PLDA model~\cite{ioffe2006probabilistic}, this probability can be expressed in the form of the normal distribution as follows:
\begin{equation}
f(\mathbf{x}~|~\tau^{y}) = \mathcal{N}(u~|~\frac{\Psi}{2\Psi+\text{I}}~u^y, \frac{\Psi}{2\Psi+\text{I}}+\text{I}),
\end{equation}
where $u$ is the image $\mathbf{x}$ transformed by the shift vector $\mathbf{m}$ and rotation and scaling matrix $A$ in the PLDA layer. Likewise, the image $\tau^y$ is transformed to $u^y$. $\Psi$ is another parameter in the learned PLDA layer. To incorporate the PLDA layer in the user model (explainee model), we train a ResNet-18 where the final layer is replaced with the PLDA layer using the user annotations as training labels. Utilizing the trained user model, we can compute $f(\mathbf{x}|\tau^{y})$, enabling the selection of images based on the ranking of this term.

\subsection{Active Learning Baselines}
Our paper incorporates baselines derived from active learning. These baselines provide different selection strategies, which are used to highlight the effectiveness of our proposed \query{}. Expected Gradient Length~\cite{settles2007multiple} (EGL) is calculated as follows:
\begin{equation}
     x_{\text{EGL}} = \argmax_{x} \summation{K}{i} f_\theta(y_i~|~\mathbf{x}, \mathbf{e})~ \Vert \nabla~l_{\theta}(\mathcal{L} \cup \langle \mathbf{x}, \mathbf{e}, y_i\rangle) \Vert,
\end{equation}
where $f_\theta(\cdot)$ denotes the trained user model in our case with parameters $\theta$. To include $\mathbf{e}$ in the input, we use the explanation $\mathbf{e}$ as the weighted mask in the same manner as proposed in the section ``Selection Strategy". $\mathcal{L}$ is the objective function for the model training, which is the cross-entropy loss. Let $\nabla~l_\theta(\mathcal{L})$ be the gradient of the objective function with respect to $\theta$. The Euclidean norm of the objective function, $\Vert \nabla~l_\theta(\mathcal{L})\Vert$ should be nearly zero since the model converged in the last round of training~\cite{settles2007multiple}. Therefore, $x_{EGL}$ can be simplified as:
\begin{equation}
     x_{\text{EGL}} = \argmax_{\mathbf{x}} \summation{K}{i} f_\theta(y_i~|~\mathbf{x}, \mathbf{e})~ \Vert \nabla~l_{\theta}(\langle \mathbf{x}, \mathbf{e}, y_i\rangle) \Vert.
\end{equation}
We extend EGL with the belief shift of the EGL when considering only $\mathbf{x}$ in the input (denoted as EGL-Shift). With EGL-Shift, we aim to alleviate the influence of an image itself on the training gradient but emphasize
the impact of explanations. Concretely, we compute the EGL-Shift as follows:
\begin{align}
         x_{\text{EGL-Shift}} = \argmax_{\mathbf{x}}~(&\summation{K}{i} f_\theta(y_i~|~\mathbf{x}, \mathbf{e})~ \Vert \nabla~l_{\theta}(\langle \mathbf{x}, \mathbf{e}, y_i\rangle) \Vert \\ \nonumber
         -&  \summation{K}{i} f_\theta(y_i~|~\mathbf{x})~ \Vert \nabla~l_{\theta}(\langle \mathbf{x}, y_i\rangle) \Vert).
\end{align}

Density-Weighted Method (DWM)~\cite{settles2008analysis} can be combined with a base selection strategy, such as EGL. Particularly, it chooses data points that are uncertain but also representative of the underlying distribution of the input data. The distribution for a data point is estimated using the similarity between this point and other points in the dataset. Specifically, DWM is conducted as follows:
\begin{equation}
    x_{\text{DWM}} = \argmax_{\mathbf{x}}~\phi_{A}(\mathbf{x}) \cdot (\frac{1}{U}\summation{U}{u=1}\text{sim}(\mathbf{x}, \mathbf{x}^{(u)}))^{\beta},
\end{equation}
where $\phi_{A}(\mathbf{x})$ denotes the calculation of EGL for $\mathbf{x}$. $U$ is the whole input dataset. Following the setting in~\cite{settles2008analysis}, we set $\beta$ to 1; The similarity between two images is calculated as the cosine similarity between the feature vectors of images in the latent space.

\section{Computational Infrastructure}
All experiments in this paper are conducted on the device as listed below:
\begin{table}[h]
\centering
\resizebox{.95\linewidth}{!}{
\begin{tabular}{c|c}
\toprule[1pt]
Device Attribute  & Value \\\hline 
Computing infrastructure&     GPU              \\
GPU model &    NVIDIA GeForce RTX 2080 Ti              \\
GPU number  &     1                   \\
CUDA version   &     11.3             \\
\bottomrule[1pt]
\end{tabular}
}
\caption{Computational infrastructure details.}
\label{supp-tab:device}
\end{table}

\section{User Study Details}
We present the procedure and some essential details of our human user study in this section.
\subsection{User Study Procedure}
The procedure of our user study is as follows:
\begin{enumerate}
    \item Participants complete a demographic survey, such as their experience with AI models. 
    \item Participants complete the warmup task. By doing this, participants adapt their reasoning to the simulated model, for which the examples on the following page are selected.
    \item Participants complete the experimental task. They are asked to the model's classification for 15 images.
    \item Participants complete a questionnaire to rate their subjective understanding of model explanations.
    \item Repeat Steps 2-4 on another dataset.
\end{enumerate}

Before the beginning of the experimental task (Step 3), participants are asked to choose the task that they will do. Choices are ``I will choose the label that I think is correct for the image" and ``I will choose the label that I think the model would predict". This single-choice question also serves as an attention check. By doing this, we can control whether all participants fully understand the task. All participants in our user study made the correct choice, i.e., ``choose the label that the model would predict".

\subsection{Objective Understanding Questions}
\begin{figure}
    \centering
    \includegraphics[width=\linewidth]{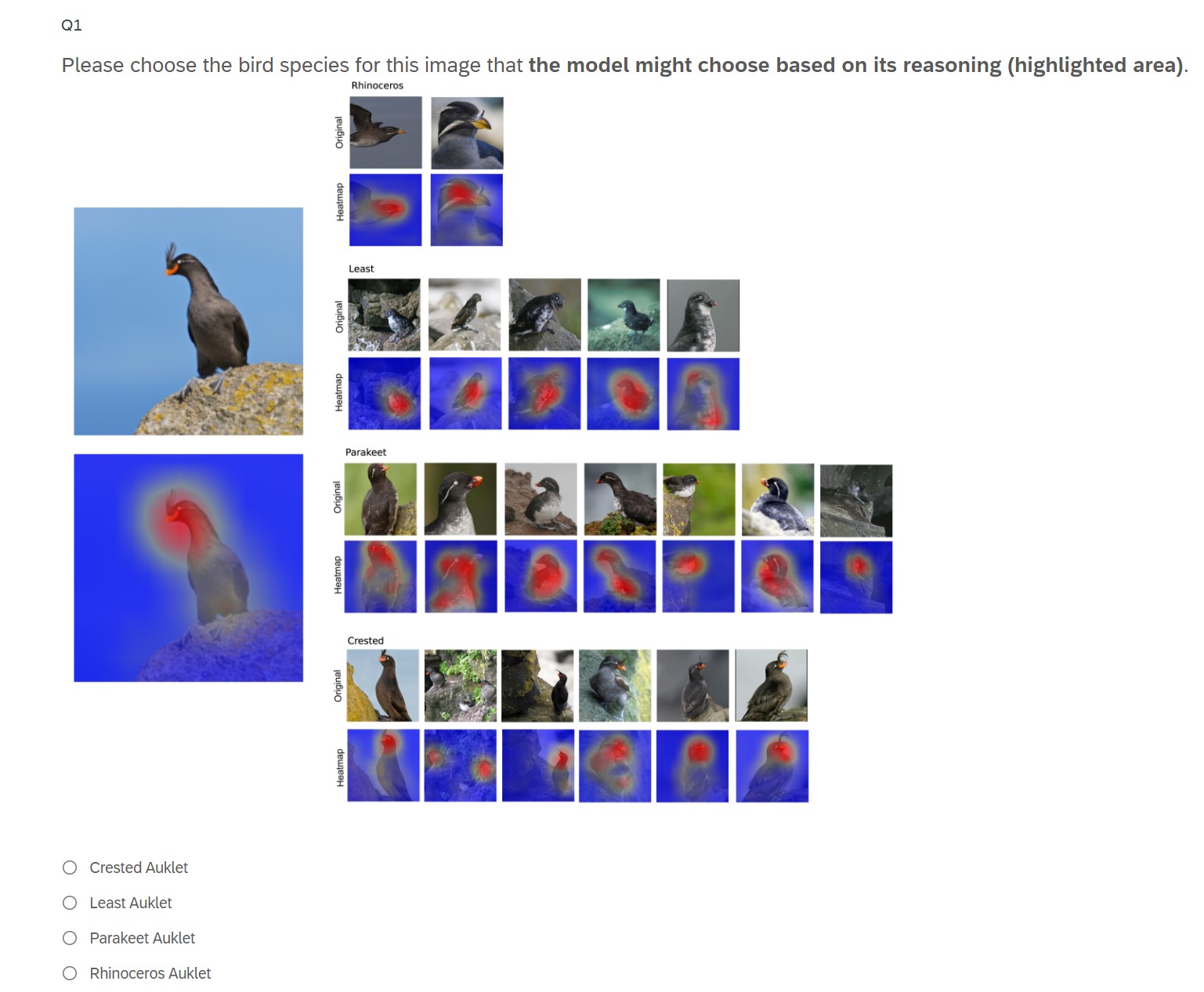}
    \caption{Question on objective understanding: participants are asked to predict the model's prediction given selected model explanations.}
    \label{fig:usr study q1}
\end{figure}

\Cref{fig:usr study q1} gives an example question used in our user study for objective understanding (simulatability). In total, there are 15 questions and they almost equally cover all four different classes. The test image is shown on the left, with the selected model explanations on the right (top 20 in the ranking according to the selection strategy). In two groups (control and experimental groups), the examples on the right are selected by different algorithms but test images on the left are the same for the two groups.

\subsection{Subjective Understanding Questions}
We use the following question for measuring subjective understanding adapted from~\cite{silva2023explainable,liao2021question}, which are answered with a 7-point Likert scale (1--Strongly Disagree; 7--Strongly Agree).

\begin{itemize}
    \item I understood the explanations within the context of this study.
    \item The explanations provided enough information for me to understand how the Machine Learning model arrived at its label. (Alternative: I would need more information to understand the explanations.)
    \item I think that most people would learn to understand the explanations very quickly. 
    \item I would like to have more examples to understand the machine's reasoning and how the machine arrived at its labeling. 
    \item The explanations were useful and helped me understand the machine's reasoning. 
    \item I believe that I could provide an explanation similar to the machine's explanation for a new image. 
\end{itemize}

\end{document}